%% file: acl2023.tex
\pdfoutput=1

\documentclass[11pt]{article}

\usepackage{ACL2023}

\usepackage{times}
\usepackage{latexsym}
\usepackage{booktabs}

\usepackage[T1]{fontenc}

\usepackage[utf8]{inputenc}

\usepackage{microtype}

\usepackage{inconsolata}
\usepackage{amsmath}
\usepackage{amssymb}
\usepackage{graphicx}
\usepackage{array}   
\usepackage{multirow}
\usepackage{makecell}
\usepackage{rotating}
\usepackage{caption}
\usepackage{subcaption}
\usepackage{enumitem}
\usepackage{bm}
\usepackage{color}
\usepackage{float,eucal}
\usepackage[ruled]{algorithm2e}
\usepackage{wrapfig}
\usepackage{footnote}

\usepackage{xcolor}
\usepackage{colortbl}  
\usepackage{makecell}

\definecolor{myblue}{RGB}{218,232,252}

\def\eg{\emph{e.g.}} 
\def\ie{\emph{i.e.}}

\newcommand{\equationref}[1]{Eq.~(\ref{#1})}
\title{LoRAPrune: Structured Pruning Meets Low-Rank \\ Para\-me\-ter-Effi\-ci\-ent Fine-Tuning}

\author{Mingyang Zhang$^{\dag\ddag}$,  ~~
  Hao Chen$^{\dag}$,  ~~
  Chunhua Shen$^{\dag\S}$,  ~~ Zhen Yang$^{\dag}$, ~~
  \textbf{Linlin Ou}$^{\ddag}$, 
  \\ {\bf Xinyi Yu}$^\ddag$\thanks{~~X. Yu is the corresponding author.}{\bf,} ~~  {\bf Bohan Zhuang}$^\dag$
  \\[0.25cm] 
  $^\dag$ Zhejiang University 
  \quad
  $^\ddag$ Zhejiang University of Technology 
  \quad 
  $^\S$ Ant Group 
}

\begin{document}
\maketitle 

\begin{abstract}
Large Language Models (LLMs), such as LLaMA and T5, have shown exceptional performance across various tasks through fine-tuning. Although low-rank adaption (LoRA) has emerged to cheaply fine-tune these LLMs on downstream tasks, their deployment is still hindered by the vast model scale and computational costs. 
Post-training model pruning offers a way to compress LLMs. However, the current pruning methods designed for LLMs are not compatible with LoRA. This is due to their utilization of unstructured pruning on LLMs, impeding the merging of LoRA weights, or their dependence on the gradients of pre-trained weights to guide pruning, which can impose significant memory overhead.
To this end, we propose LoRAPrune, a new framework that delivers an accurate structured pruned model in a highly memory-efficient manner. Specifically, we first design a LoRA-guided pruning criterion, which uses the weights and gradients of LoRA, rather than the gradients of pre-trained weights for importance estimation. We subsequently integrate this criterion into an iterative pruning process, effectively removing redundant channels and heads. 
Extensive experimental results demonstrate the superior performance of our LoRAPrune over existing approaches on the LLaMA series models.
At a 50\% compression rate, LoRAPrune demonstrates superior performance over LLM-Pruner, achieving a reduction in perplexity by 4.81 on WikiText2 and 3.46 on PTB, while also decreasing memory usage by 52.6\%.
Besides, LoRAPrune also matches semi-structural pruning across multiple LLMs, proving its wide applicability. The code is available at \url{https://github.com/aim-uofa/LoRAPrune}.
\end{abstract}
\section{Introduction}
Large Language Models (LLMs) \citep{touvron2023llama,du2022glm,frantar2023gptq} have showcased remarkable prowess, exhibiting outstanding performance across numerous tasks.
To enable LLMs to perform specific tasks, such as chat-bots \citep{du2022glm,zeng2022glm}, they are often efficiently fine-tuned on downstream datasets \citep{alpaca,leng2023chinese-vicuna} by parameter-efficient fine-tuning (PEFT) methods \citep{jia2022vpt,hu2022lora,chen2022adaptformer}, among which LoRA-based fine-tuning methods \citep{hu2022lora,luo2023towards,he2023sensitivity} have gained widespread use. 
However, the remarkable success of LLMs is accompanied by obstacles from their vast scale and substantial computational costs, making deployment exceedingly arduous \citep{frantar2023massive}. 

\begin{table}
\caption{The memory costs for pruning LLaMA-65B. ``Iter.'' indicates whether the method supports iterative pruning and  ``\#GPU" indicates the number of NVIDIA A100 (80G) GPUs required.}\label{tab:com_mem}
 \resizebox{\linewidth}{!}{%
    \begin{tabular}{c|c|c|c}
        \hline
        Method&Iter.&\#GPU&Mem.(G)\\
        \hline
        PST~\citep{li2022parameter}&$\checkmark$&3&234\\
        LLM-Pruner~\citep{ma2023llm}&$\times$&2&154\\
        LoRAPrune&$\checkmark$&1&72\\
        \hline
    \end{tabular} 
}
\end{table}
\begin{figure*}[t]
    \centering
    \includegraphics[scale=0.45]{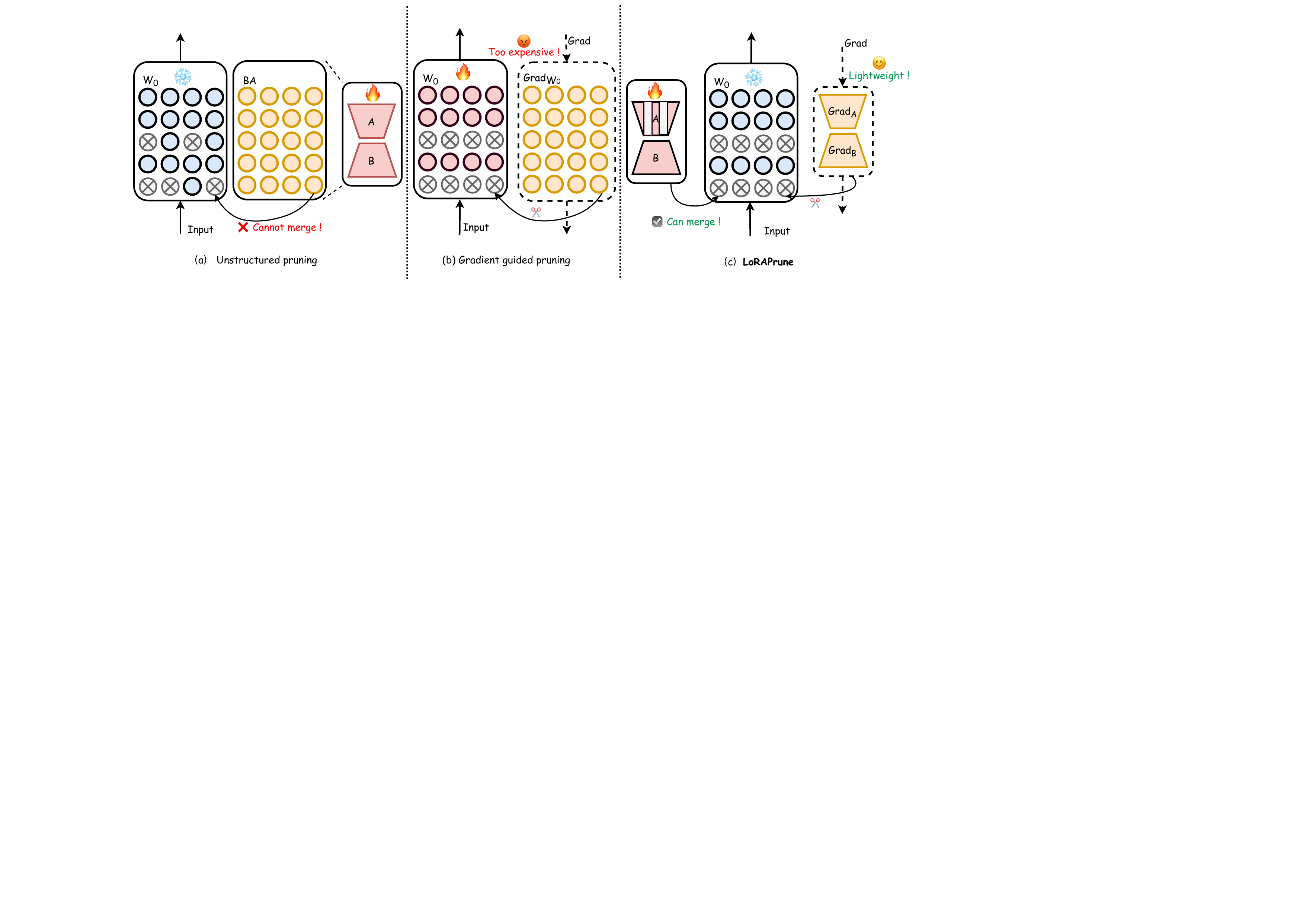}
    \caption{\label{fig:overview} 
    Comparing LoRAPrune with other pruning methods: (a) Unstructured sparse model cannot directly merge LoRA weights, which is computationally inefficient. (b) Gradient-guided pruning requires the gradients of the pre-trained weights, which is memory-intensive. (c) LoRAPrune only needs the gradients of LoRA weights and can seamlessly merge LoRA weights into pre-trained weights, which is efficient in both memory and computation.}
\end{figure*}
\noindent Neural network pruning \citep{li2017pruning, molchanov2016pruning}, a 
prevailing
technique for model compression, can significantly reduce the model size and complexity. 
Recently, the post-training pruning literature, such as SparseGPT \citep{frantar2023massive} and WANDA \citep{sun2023simple}, have achieved high-performance unstructured sparse LLMs. However, unstructured sparse models face two critical issues: \emph{1)} 
\emph{Unstructured sparse models are hard to obtain direct inference speedup}. They often require specialized hardware support to achieve satisfying acceleration benefits, which leads to unstructured pruning not benefiting legacy off-the-shelf platforms, 
\eg,
CPUs, DSPs, and GPUs \citep{Fang_2023_CVPR,you2023vitcod,zhou2022transpim}.  
\emph{2)} \emph{Unstructured sparse models are not compatible with LoRA.} As shown in Figure \ref{fig:overview} (a), since the weights $\mathbf{BA}$ produced by LoRA are dense, it poses challenges when trying to merge $\mathbf{BA}$ into the unstructured sparse weights. For instance, LoRA without merging increases inference time by nearly 54\% (see Table \ref{tab:speedup}), diminishing the benefits of pruning. One potential solution is to perform fine-tuning using LoRA on downstream tasks first and then carry out post-training pruning. However, separating tuning and pruning can lead to sub-optimal results \citep{molchanov2019importance,sanh2020movement}. To tackle this challenge, PST \citep{li2022parameter} combines unstructured pruning with efficient fine-tuning, which simultaneously prunes LoRA and pre-trained weights. This method ensures a seamless merge of LoRA weights and avoids additional computational overhead that comes from LoRA. However, unstructured pruning of LoRA necessitates computing $\mathbf{BA}$ first and then doing Hadamard product with a binary mask $\mathbf{M}$, which results in significant memory overhead (see Table \ref{tab:com_mem}) since $\mathbf{BA}$ and $\mathbf{M}$ share the same shape with pre-trained weights. For instance, when pruning LLaMA-65b, the intermediate variables necessitate the storage capacity of three NVIDIA A100 (80G) GPUs. This poses a significant memory challenge when adapting PST to LLMs. Instead, structured pruning can mitigate this issue since we can directly prune the structured weights of $\mathbf{A}$ in LoRA without storing $\mathbf{BA}$. Therefore, it is significant to combine LoRA with structured pruning to achieve simultaneous PEFT and direct acceleration on general hardware platforms with high performance.

\begin{figure*}[t]
\begin{center}
\includegraphics[scale=0.605]{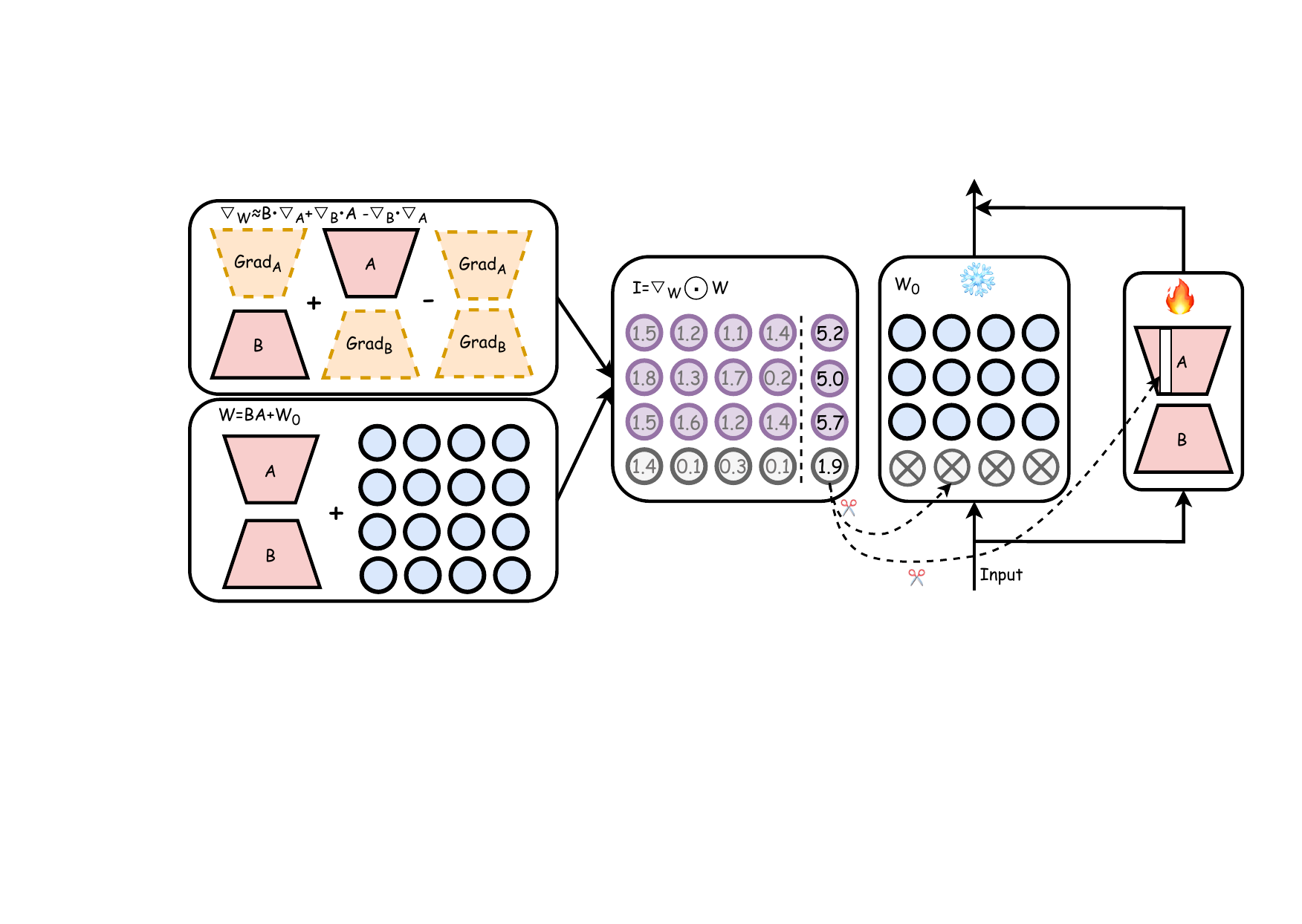}
\end{center}
\caption{
The pruning process for the LoRA-guided criterion involves utilizing the LoRA matrices $\mathbf{A}$, $\mathbf{B}$ and their respective gradients $\nabla_\mathbf{A}$, $\nabla_\mathbf{B}$ to compute the importance $\mathbf{I}$. Subsequently, weight importance (\textcolor[RGB]{116,114,114}{gray number}) with the same group are aggregated to the group importance (\textcolor{black}{black number}) and the groups with low scores are removed. 
}
\label{fig:criterion}
\end{figure*}

\noindent To this end, we propose a unified framework for LoRA and structured pruning, named LoRAPrune. 
As shown in Figure \ref{fig:overview} (c),
LoRAPrune not only prunes the structured weights (\eg, heads, channels) from the pre-trained model weights $\mathbf{W}_0$ but also trims the corresponding weights in LoRA weight $\mathbf{A}$ without computing $\mathbf{BA}$ first. 
Consequently, after pruning and fine-tuning, the weights of LoRA can be \emph{seamlessly} merged with the pre-trained weights, ensuring that no additional computations are needed during inference. 
To identify weight connections of structural importance, the criterion used in the structured pruning methods \citep{ma2023llm,molchanov2019importance,molchanov2016pruning} is often estimated by gradients or its variants, as shown in Figure \ref{fig:overview} (b).
However, LoRA typically requires frozen pre-trained weights without computing their gradients, 
thus 
pruning approaches that rely on gradients of the pre-trained weights cannot be directly applied. To \emph{efficiently} estimate the importance of pre-trained weights, LoRAPrune introduces a 
novel criterion that exclusively utilizes the gradients of LoRA.
In contrast to the vanilla gradient-guided pruning method, LoRAPrune leverages LoRA's gradients as the approximation for the gradients of the pre-trained weights.
Based on the presented criterion, we can \emph{iteratively} perform pruning while simultaneously conducting efficient fine-tuning to restore the performance of the pruned LLMs, requiring only a small calibration set. 
Specifically, we compute the importance of every batch of data and update the importance using a moving average. Every few iterations, we remove a portion of unimportant structured weights until the desired sparsity is achieved.
Through extensive experiments on 
diverse benchmark datasets and various scales of LLMs,
we demonstrate that LoRAPrune consistently outperforms other structured pruning techniques tailored for LLMs. Furthermore, compared to the vanilla gradient-guided pruning, LoRAPrune 
significantly diminishes
memory and computational overhead, 
facilitating
efficient pruning and fine-tuning 
of
LLaMA-65b on
a single
GPU concurrently. This paper has the following key contributions:

\begin{itemize}[leftmargin=*]

\item We introduce a novel memory-efficient pruning
criterion tailored for LLMs, termed the LoRA-guided criterion, 
which
seamlessly 
integrates
with LoRA.
Leveraging
the gradients of LoRA, we can efficiently approximate the importance of pre-trained weights without needing to compute their gradients.

\item 
As we can efficiently approximate gradients and update weights using LoRA, LoRAPrune facilitates iterative structured pruning, resulting in precise small models. Our framework ensures both high memory efficiency during pruning and incurs efficient inference.

\item Pruning experiments conducted on the LLaMA models demonstrate that LoRAPrune can efficiently perform structured pruning with up to 65 billion weights on a single GPU. Furthermore, the pruning results achieved by LoRAPrune significantly surpass other pruning methods.  For example, against LLM-Pruner, LoRAPrune uses only 52.6\% of the memory yet scores lower perplexities by 4.81 on WikiText2 and 3.46 on PTB. LoRAPrune also matches semi-structural pruning performance across various LLMs, proving its broad applicability.
\end{itemize}

\section{Related Work}
\textbf{Parameter-efficient fine-tuning.} 
PEFT methods \citep{jia2022vpt,wu2022generative,chen2022adaptformer,hu2022lora,luo2023towards,he2023sensitivity} have received increasing attention from both academia and industry. Among them, 
LoRA \citep{hu2022lora} proposes injecting trainable low-rank decomposition matrices into each layer which can be merged into the pre-trained weights, avoiding extra computation in inference. Since inference efficiency, many methods based on LoRA have emerged. For instance, LongLoRA \citep{chen2023longlora} improves upon LoRA, enabling efficient fine-tuning of LLMs on long contexts. AnimateDiff \citep{guo2023animatediff} obtains a personalized generator by inserting LoRA into the frozen text-to-image model. Quantizing the pre-trained weights into 4-bit, QLoRA \citep{dettmers2023qlora} employs LoRA for fine-tuning LLMs in downstream tasks while maintaining efficient memory usage. Therefore, LoRA is indispensable for fine-tuning LLMs. Our method seamlessly integrates LoRA and pruning, making it easily extensible to other PEFT methods based on LoRA.

\noindent\textbf{Neural network pruning.}
Removing unimportant weights from LLMs to reduce memory
and the computational cost of deployment has become a common approach for model compression. Unstructured pruning \citep{dong2017learning,lee2018snip,wang2020picking,sun2023simple,frantar2023massive,li2022parameter} can obtain highly compressed models by directly pruning neurons, which also causes unstructured sparsity and hard deployment. In contrast, structured pruning \citep{ma2023llm,xia2023sheared,guo2023compresso} directly discards the whole grouped parameters (\eg heads, channels) and leaves a model with deploy-friendly structures. However, structured pruning models require extensive finetuning to regain their performance levels. For example, \citet{xia2023sheared} utilized 50B tokens sampled for continued pretraining of their pruned model, a process that proves to be prohibitively expensive in terms of hardware resources. In contrast, our approach leverages structured pruning, enabling direct inference acceleration while maintaining training expenses at an acceptable level.

\noindent\textbf{Pruning criterion.} Determining the importance of weights in a network is still an open question \citep{blalock2020state}. A common approach to model pruning is to use parameter magnitude \citep{li2018optimization,lee2020layer,elesedy2020lottery,han2015learning,li2017pruning} as a criterion. 
However, the small weights can still have a significant impact on the model output due to the complex structure of neural networks, while large weights may not be as important. Many methods 
\citep{sanh2020movement,yu2022width,zhang2022platon,lee2018snip, yu2022combinatorial,wang2020picking,lecun1989optimal,hassibi1993optimal}
employ Taylor expansion to approximate the errors introduced by pruning and use this as the criterion for importance estimation. To avoid computing the Hessian matrix \citep{hassibi1993optimal} or Hessian inverse \citep{lecun1989optimal} in Taylor expansion, \cite{molchanov2016pruning,molchanov2019importance} only use the first-order term in Taylor expansion. Furthermore, LLM-Pruner \citep{ma2023llm} similarly utilizes the first-order
expansion for pruning and extends the pruning technique to LLMs. However, the first-order term in Taylor expansion still requires gradients of the pre-trained weights. As shown in Table \ref{tab:com_mem}, computing and storing the gradients of pre-trained weights significantly increases the pruning cost. To avoid computing gradients of pre-trained weights, PST \citep{li2022parameter} learns the gradients of pre-trained weights by an extra low-rank matrix, which is motivated by LoRA. Nevertheless, PST conducts unstructured pruning and needs to compute a substantial mask with the equivalent shape of pre-trained weights in each forward pass, which is memory-intensive and hard to be adapted to LLMs. Different from LLM-Pruner \citep{ma2023llm} and PST \citep{li2022parameter}, our criterion only relies on LoRA's gradients and does not require expensive mask computation, making it memory-efficient. 

\section{Method}

\subsection{Preliminary}
\label{preliminary}
We define the notation 
used in this paper. 
\textbf{Bold} letters represent matrices and vectors.
Lower-case 
letters indicate scalars. ``Subscripts'' identify the index of elements within a matrix, and ``superscripts'' indicate the layer index in a network.

\noindent \textbf{Low-rank adaptation.} 
To efficiently fine-tune LLMs, low-rank adapter LoRA \citep{hu2022lora} constrains the update of model parameters to maintain a low intrinsic rank.
During fine-tuning, the pre-trained weights remain frozen, abstaining from gradient computation, while the inserted LoRA is kept trainable.
Given two low-rank matrices $\mathbf{A} \in \mathbb R^{r\times k}$ and $\mathbf{B} \in \mathbb R^{d \times r}$ ($r  \ll  \min(d,k)$), 
the update of a linear module can be written as
\begin{equation}
\label{eq:lora}
    \mathbf{z} = \mathbf{\mathbf{x}}\mathbf{W}_0 + \mathbf{\mathbf{x}}\mathbf{BA},
\end{equation}
where $\mathbf{W}_0 \in \mathbb R^{d\times k}$, $\mathbf{z} \in \mathbb R^{n\times k}$ and $\mathbf{\mathbf{x}} \in \mathbb R^{n\times d}$ denote the pre-trained weights, outputs and inputs, respectively. After adaption, the new weights $\mathbf{W}$ can be re-parameterized as $\mathbf{W}=\mathbf{W}_0 + \mathbf{BA}$.

\noindent \textbf{Pruning with Taylor expansion.} In vanilla pruning approaches \citep{molchanov2016pruning,molchanov2019importance}, the importance of a weight $\mathbf{W}_{i,j} \in \mathbf{W}_0$ can be quantified by measuring the impact of its removal on the loss. For an input $\mathbf{x}$ and the ground-truth prediction $\mathbf{y}$,
the induced error of $\mathbf{W}_{i,j}$ can be given as:
\begin{equation}
\label{importance_w0}
    \mathbf{I}_{i,j}\ =[\mathcal{L}(\mathbf{x}, \mathbf{y}, \mathbf{W}_0) - \mathcal{L}(\mathbf{x}, \mathbf{y}, \mathbf{W}_0|\mathbf{W}_{i,j} = 0)]^2.
\end{equation}
Computing $\mathbf{I}_{i,j}$ for each weight is computationally expensive. Following \cite{molchanov2019importance}, we can use first-order Taylor expansion to approximate the importance $\mathbf{\hat{I}}_{i,j}$ by:
\begin{equation}
\label{importance_w}
    \mathbf{\hat{I}}_{i,j}\ = (\frac{\partial \mathcal{L}}{\partial \mathbf{W}_{i,j}} \mathbf{W}_{i,j})^2.
\end{equation}

\noindent \textbf{Dependency-aware structured pruning.} In structured pruning, it is crucial to consider that pruned neurons can exhibit dependencies with other neurons due to their interconnected nature. 
The dependencies of weights are illustrated in Figure \ref{fig:dependency}. We organize the connected weights as a group and estimate the group importance by accumulating the weight importance within the same group. Formally, the importance for the $g$-th group can be expressed as
\begin{equation}
    \label{eq:group_imp}
    \bm{\mathcal{\hat{G}}}_g = \sum_{\mathbf{W}_{i,j}\in \mathbb{G}} \mathbf{\hat{I}}_{i,j} \ ,
\end{equation}
where $\bm{\mathcal{\hat{G}}} \in \mathbb{R}^{1\times G}$ represents the importance of groups, $\mathbb{G}$ denotes a set of weights within a group
and $G$ is the candidate group number in a layer. 

\begin{algorithm}[t]
\SetKwInOut{Input}{Require}
\SetKwInOut{Output}{Output}
\SetAlgoLined
    \caption{Progressive pruning with LoRA-guided criterion}
    \label{alg:LoRAPrune}
    \Input{Calibration data $\mathcal{D}$; Pre-trained weights $\mathbf{W}_0$; Randomly initialized low-rank matrices $\mathbf{A}$ and $\mathbf{B}$; Loss function $\mathcal{L}$; Target sparsity level $S$; Fine-tuning iterations $T$.}
    \Output{Trained low-rank adaption $\mathbf{A}$ and $\mathbf{B}$; Binary mask $\mathbf{M}$.}
    $\bm{\bm{\mathcal{\bar{G}}}}^l_{g}$ $\leftarrow$ 0, $\mathbf{M}^l_{g}$ $\leftarrow$ 1 for $\forall l,\forall g$; \tcp{Initialization for masks and group importance} 
    $s$ $\leftarrow$ 0; \tcp{Initialize sparsity level}
    \For{$t \in [1, \dots, T]$}{
        Clear gradient; \\
        Forward and backward \text{via ~Eq. (\ref{forward})}; \\
        Update $\mathbf{A}$ and $\mathbf{B}$ via AdamW; \\
        Calculate $\mathbf{\hat{I}}|_{t}$ \text{via Eq. (\ref{eq:final_imp})}; \\
        Calculate $\bm{\mathcal{\hat{G}}}|_{t}$ \text{via Eq. (\ref{eq:group_imp})}; \\
        Calculate $\bm{\mathcal{\bar{G}}}|_{t}$ \text{via ~Eq. (\ref{moving_average})}; \\
        \For{$l \in [1, \dots, L]$}{
        $p$ $\leftarrow$ $\rm{SortDescending}(\bm{\mathcal{\bar{G}}})_s$; \tcp{Set threshold} 
        $\mathbf{M}^l_{g}$ $\leftarrow$ 0 where 
        $\bm{\mathcal{\bar{G}}}^l_{g} \leq p$, and $g \in\{1, \dots, G\}$} \tcp{Remove unimportant groups} 
        
        Progressively increase $s$ until $||\mathbf{M}||_0 > S$;}
\end{algorithm}

\begin{table*}[t]
    \centering
    \caption{Zero-shot performance of the compressed LLaMA models fine-tuned on the LaMini dataset. 
    We evaluate WikiText2 and PTB on perplexity with 2048-token segments. The average accuracy is calculated among seven classification datasets. \textbf{Bold} denotes the best performance at the same compression rate.
    $^{\star}$ denotes the results obtained by our reproduction. 
    } \label{tab:LLaMA_result}
    \resizebox{\linewidth}{!}{
    \begin{tabular}{l|l|cc|c|cccccccc}
    \hline
        Pruning Ratio & Method & WikiText2$\downarrow$ & PTB$\downarrow$ & MMLU (5-shot)& OBQA&ARC-e &WinoGrande &ARC-c &PIQA & HellaSwag & Average$\uparrow$ \\
        \hline
        \hline
        \multirow{1}{*}{Ratio = 0\%} & LLaMA-7B~\citep{touvron2023llama} & 5.69 & 8.93 & 37.10 & 42.40 & 67.45 & 67.01 & 67.45 & 78.35 & 72.99 & 65.34 \\
        \hline
        \hline
        \multirow{6}{*}{\parbox{2.0cm}{Ratio = 20\%}} 
        &Magnitude $^{\star}$ & 9.06 & 13.80 & 27.84 & 35.80&	65.36&	61.33&	38.74&	74.87&	63.90&	56.67 \\
        &WANDA$^{\star}$ \citep{sun2023simple}& 8.64 & 12.66 & 28.35 &35.26&	68.96&	64.01&	38.46&	74.80&	52.63&	58.68 \\
        & LLM-Pruner$^{\star}$  \citep{ma2023llm}& 8.14 &  12.38 & 33.67 &\textbf{38.8}&	\textbf{70.62}&	65.82&	40.7&	77.37&	66.6&	62.36 \\
        & Compresso \citep{guo2023compresso} & - &  - & 31.90 &36.4&	68.64&	\textbf{67.80}&	37.97&	75.46&	53.44&	59.82 \\
        &\cellcolor{myblue}LoRAPrune-8bit (Ours)& \cellcolor{myblue}7.70 & \cellcolor{myblue}11.91 & \cellcolor{myblue}36.45 & \cellcolor{myblue}38.1&	\cellcolor{myblue}70.25&	\cellcolor{myblue}65.93&	\cellcolor{myblue}41.43&	\cellcolor{myblue}77.10&	\cellcolor{myblue}\textbf{68.90}&	\cellcolor{myblue}60.29 \\
        & \cellcolor{myblue}LoRAPrune (Ours)& \cellcolor{myblue}\textbf{7.63} & \cellcolor{myblue}\textbf{11.87}  & \cellcolor{myblue}\textbf{36.81} & \cellcolor{myblue}38.6&	\cellcolor{myblue}70.20&	\cellcolor{myblue}66.77&	\cellcolor{myblue}\textbf{41.89}&	\cellcolor{myblue}\textbf{77.48}&	\cellcolor{myblue}68.64&	\cellcolor{myblue}\textbf{62.70} \\
        \hline
        \hline
        \multirow{6}{*}{\parbox{2.0cm}{Ratio = 30\%}}
       &Magnitude $^{\star}$ & 11.38 & 16.90 & 26.38 & 33.67& 65.58& 60.79& 37.47& 73.15&60.35&	55.16 \\
        &WANDA$^{\star}$ \citep{sun2023simple}& 10.10 & 15.83 & 27.90 &34.90&	65.06&	61.16&	39.44&	74.38&	60.84&	55.96 \\
        & LLM-Pruner$^{\star}$  \citep{ma2023llm}& 9.36 &  13.82 &30.67 &34.86&	66.2&	63.85&	40.55&	75.60&	65.12&	57.70  \\
        & Compresso \citep{guo2023compresso} & - &  - & 27.68 &29.8&	66.23&	\textbf{64.80}&	37.2&	75.63&	49.16&	53.79 \\
        & \cellcolor{myblue}LoRAPrune-8bit (Ours)& \cellcolor{myblue}8.83 & \cellcolor{myblue}\textbf{13.30}  & \cellcolor{myblue}33.36 &	\cellcolor{myblue}\textbf{36.40}&	\cellcolor{myblue}69.48&	\cellcolor{myblue}62.31&	\cellcolor{myblue}\textbf{41.93}&	\cellcolor{myblue}77.40&	\cellcolor{myblue}65.91&	\cellcolor{myblue}58.90 \\
        & \cellcolor{myblue}LoRAPrune (Ours)& \cellcolor{myblue}\textbf{8.79} & \cellcolor{myblue}13.33  & \cellcolor{myblue}\textbf{33.60} &	\cellcolor{myblue}36.20&	\cellcolor{myblue}69.61& \cellcolor{myblue}62.75&	\cellcolor{myblue}41.21&	\cellcolor{myblue}\textbf{77.48}&	\cellcolor{myblue}\textbf{66.68}&	\cellcolor{myblue}\textbf{58.98} \\
        \hline
        \hline
        \multirow{5}{*}{\parbox{2.0cm}{Ratio = 50\%}}
       &Magnitude $^{\star}$ & 18.36 & 23.88 &21.84 & 30.26&53.61&55.86&36.98&67.10&53.10&49.48 \\
        &WANDA$^{\star}$ \citep{sun2023simple}& 17.38 & 21.34 & 24.15& 28.78&52.68&55.98&34.20&70.38&54.12&49.35 \\
        & LLM-Pruner$^{\star}$  \citep{ma2023llm}&16.41 & 20.85  & 25.60 &	33.12&	55.36& 56.12&	34.98&	73.25&	58.60&	51.90 \\
        & \cellcolor{myblue}LoRAPrune-8bit (Ours)& \cellcolor{myblue}11.65 & \cellcolor{myblue}17.41  & \cellcolor{myblue}27.71 &	\cellcolor{myblue}35.30&	\cellcolor{myblue}\textbf{60.54}& \cellcolor{myblue}56.13&	\cellcolor{myblue}\textbf{40.58}&	\cellcolor{myblue}74.89&	\cellcolor{myblue}59.86&	\cellcolor{myblue}54.55 \\
        & \cellcolor{myblue}LoRAPrune (Ours) &\cellcolor{myblue}\textbf{11.60} & \cellcolor{myblue}\textbf{17.39}  & \cellcolor{myblue}\textbf{27.84} &	\cellcolor{myblue}\textbf{35.80} & \cellcolor{myblue}60.38 &\cellcolor{myblue}\textbf{56.97}& \cellcolor{myblue}40.12&	\cellcolor{myblue}\textbf{75.39}&	\cellcolor{myblue}\textbf{60.21}&	\cellcolor{myblue}\textbf{54.81} \\
        \hline
    \end{tabular}
    }
\end{table*}
\begin{table*}
\caption{Runtime results of the structured pruned LLMs. }\label{tab:speedup}
\centering
\scalebox{0.806}{
    \begin{tabular}{c|c|c|c|c}
        \hline
        Model & Unmerged time (s) $\downarrow$& Merged time (s) $\downarrow$ & Perplexity $\downarrow$ & Ratio (\%)\\
        \hline
        \multirow{3}{*}{\parbox{1.8cm}{LLaMA-7B}}& 0.184(+0.0\%)&  0.105(+0.0\%)& 5.69& 0\\
         &  0.120(\textcolor{green}{-34.8\%})&  0.079(\textcolor{green}{-24.7\%})&7.63 &20\\
        &  0.089(\textcolor{green}{-51.6\%})&  0.053(\textcolor{green}{-49.5\%})&11.60&50\\
        \hline
    \end{tabular}
    }
\end{table*}

\subsection{Pruning with Low-rank Adaption}
\textbf{Motivation.} To achieve highly-compressed LLMs, it is essential to accurately evaluate the importance of pre-trained weights. A key approach is to utilize the criteria in \equationref{importance_w} for this evaluation. However, obtaining the gradient of $\mathbf{W}_0$ in a LLM is difficult since it requires a lot of computing power and storage space. Fine-tuning LLMs with LoRA is becoming prevalent \citep{alpaca,leng2023chinese-vicuna}. During LoRA fine-tuning, only the gradients of LoRA's weights are computed, yielding remarkable computation and memory efficiency. Therefore, can we rely solely on the weights and gradients of LoRA to accurately estimate the importance of pre-trained weights?

\noindent \textbf{LoRA-guided criterion.} 
In this work, we discuss how to estimate the importance of $\mathbf{W}_0$ by inserting the learnable matrices $\mathbf{A}$ and $\mathbf{B}$ in the downstream task adaption.

\noindent The core idea lies in setting the element $(\mathbf{BA})_{{ij}} = -\mathbf{W}_{{ij}}$ if the element $\mathbf{W}_{{ij}} \in \mathbf{W}_0$ is removed. The importance of each parameter in  \equationref{importance_w0} can be reformulated as follows
\begin{equation}
\label{importance_ab}
    \mathbf{I}_{i,j}\ = [\mathcal{L}(\mathbf{x}, \mathbf{y}, \mathbf{W}) - \mathcal{L}(\mathbf{x}, \mathbf{y}, \mathbf{W}|(\mathbf{BA})_{i,j} = -\mathbf{W}_{i,j}]^2.
\end{equation}
Exploiting the first-order Taylor expansion with $(\mathbf{BA})_{i,j}=-\mathbf{W}_{i,j}$ to approximate  \equationref{importance_ab}, the estimated importance $\mathbf{\hat{I}}_{i,j}$ of parameter $\mathbf{W}_{i,j}$ can be represented by
\begin{equation}
\label{importance_score_par}
    \mathbf{\hat{I}}_{i,j} = [\frac{\partial \mathcal{L}}{\partial(\mathbf{BA})_{i,j}}((\mathbf{BA})_{i,j} + \mathbf{W}_{i,j})]^2.
\end{equation}
However, as shown in \equationref{eq:lora}, the LoRA computation sequence involves first multiplying by $\mathbf{B}$ and then by $\mathbf{A}$, which means that $\mathbf{BA}$ cannot be obtained during the forward and backward pass. Besides, preserving $\frac{\partial \mathcal{L}}{\partial(\mathbf{BA})_{i,j}}$ still entails the same level of complexity as $\frac{\partial \mathcal{L}}{\partial \mathbf{W}_{i,j}}$ since $\mathbf{BA}$ shares the same shape of $\mathbf{W}_0$.

\noindent Here, we only save and use the gradients of two low-rank matrices $\mathbf{A}$ and $\mathbf{B}$ to approximate $\frac{\partial \mathcal{L}}{\partial(\mathbf{BA})}$. We can rely on the gradient update that $(\mathbf{BA})_{i,j}|_{t} = (\mathbf{BA})_{i,j}|_{t-1} - \eta\frac{\partial \mathcal{L}}{\partial(\mathbf{BA})_{i,j}}$ to estimate the gradient, where $(\mathbf{BA})_{i,j}|_{t}$ and $(\mathbf{BA})_{i,j}|_{t-1}$ represents the $(\mathbf{BA})_{i,j}$ in $t$-th and $(t-1)$-th step, respectively. 
Apparently, $\eta \frac{\partial \mathcal{L}}{\partial(\mathbf{BA})_{i,j}}$ is equal to the change of $\mathbf{BA}$, which can be written as 
\begin{equation}
\label{grad_estimate}
    \eta \frac{\partial \mathcal{L}}{\partial(\mathbf{BA})_{i,j}} = [(\mathbf{BA})_{i,j}|_{t-1} - (\mathbf{BA})_{i,j}|_{t}].
\end{equation}
Here, $(\mathbf{BA})_{i,j}|_{t} = \mathbf{B}_{i,:}|_{t}\mathbf{A}_{:,j}|_{t}$ is generated by
the multiplication of the $i$-th row of $\mathbf{B}|_{t}$ with the $j$-th column of $\mathbf{A}|_{t}$. Using the above assumption, we can also estimate 
$\eta \frac{\partial \mathcal{L}}{\partial \mathbf{A}_{:,j}} = \mathbf{A}_{:,j}|_{t-1} - \mathbf{A}_{:,j}|_{t}$ 
and 
$\eta \frac{\partial \mathcal{L}}{\partial \mathbf{B}_{i,:}} = \mathbf{B}_{i,:}|_{t-1} - \mathbf{B}_{i,:}|_{t}$, 
respectively. Further, we can obtain that $\mathbf{A}_{:,j}|_{t} = \mathbf{A}_{:,j}|_{t-1} - \eta \frac{\partial \mathcal{L}}{\partial \mathbf{A}_{:,j}}$ and $\mathbf{B}_{i,:}|_{t} = \mathbf{B}_{i,:}|_{t-1} - \eta \frac{\partial \mathcal{L}}{\partial \mathbf{B}_{i,:}}$.  Subsequently, we can calculate
\begin{equation}
\label{eq:ba}
\begin{split}
        (\mathbf{BA})_{i,j}|_{t} &=\mathbf{B}_{i,:}|_{t}\mathbf{A}_{:,j}|_{t} \\
        &=(\mathbf{B}_{i,:}|_{t-1} - \eta \frac{\partial \mathcal{L}}{\partial \mathbf{B}_{i,:}})(\mathbf{A}_{:,j}|_{t-1} - \eta \frac{\partial \mathcal{L}}{\partial \mathbf{A}_{:,j}})\\
\end{split}
\end{equation}
Substitute $(\mathbf{BA})_{i,j}|_t $ in \equationref{eq:ba}\ to \equationref{grad_estimate} and obtain
\begin{equation}
\label{grad_final_par}
\begin{split}
    \frac{\partial \mathcal{L}}{\partial(\mathbf{BA})_{i,j}} =& [ \frac{\partial \mathcal{L}}{\partial \mathbf{B}_{i,:}}\mathbf{A}_{:,j}|_{t-1} + \mathbf{B}_{i,:}|_{t-1}\frac{\partial \mathcal{L}}{\partial \mathbf{A}_{:,j}} \\
    &- \eta \frac{\partial \mathcal{L}}{\partial \mathbf{B}_{i,:}}\frac{\partial \mathcal{L}}{\partial \mathbf{A}_{:,j}}]. 
\end{split}
\end{equation}
For simplicity, we set the learning rate $\eta = 1$. Substitute \equationref{grad_final_par} to \equationref{importance_score_par}, we can estimate the importance in a gradient-based manner
$\mathbf{\hat{I}}_{i,j} = [(\frac{\partial \mathcal{L}}{\partial \mathbf{B}_{i,:}}\mathbf{A}_{:,j} + \mathbf{B}_{i,:}\frac{\partial \mathcal{L}}{\partial \mathbf{A}_{:,j}} - \frac{\partial \mathcal{L}}{\partial \mathbf{B}_{i,:}}\frac{\partial \mathcal{L}}{\partial \mathbf{A}_{:,j}})(\mathbf{W}_{i,j} + (\mathbf{BA})_{i,j})]^2. $
\begin{equation}
    \label{eq:final_imp}
    \begin{split}
        \mathbf{\hat{I}}_{i,j} = &[(\frac{\partial \mathcal{L}}{\partial \mathbf{B}_{i,:}}\mathbf{A}_{:,j} + \mathbf{B}_{i,:}\frac{\partial \mathcal{L}}{\partial \mathbf{A}_{:,j}} - \frac{\partial \mathcal{L}}{\partial \mathbf{B}_{i,:}}\frac{\partial \mathcal{L}}{\partial \mathbf{A}_{:,j}}) \\
        &(\mathbf{W}_{i,j} + (\mathbf{BA})_{i,j})]^2.
    \end{split}
\end{equation}
As shown in Figure \ref{fig:criterion}, the LoRA-guided criterion only needs to compute the gradients of $\mathbf{A}$ and $\mathbf{B}$ with the approximation in \equationref{eq:final_imp}, which saves memory and computation compared with the gradients of pre-trained weights $\mathbf{W}_0$. 

\noindent \textbf{Progressive pruning.} To efficiently obtain group importance for structured pruning, we can substitute \equationref{eq:final_imp} into \equationref{eq:group_imp}. 
However, estimating importance and pruning weights with a single batch of data can lead to significant bias and performance loss. To mitigate this, we apply moving average to evaluate group importance $\bm{\mathcal{G}}$ and incrementally prune less critical groups. Specifically, the group importance at $t$-th iteration is computed as follows: 
\begin{equation}
\label{moving_average}
\mathbf{\bm{\mathcal{\bar{G}}}}|_{t} = \lambda\bm{\mathcal{{\bar{G}}}}|_{t-1} + (1-\lambda)\bm{\mathcal{\hat{G}}}|_{t}.
\end{equation}
Here, $\bm{\mathcal{\bar{G}}}|_{t}$ denotes the group importance scores calculated by \equationref{eq:final_imp} and \equationref{eq:group_imp} at the $t$-th iteration, and $\lambda \in [0, 1]$ balances the importance between historical and current statistics.

\begin{figure*}[t]
\begin{center}
\includegraphics[scale=0.22]{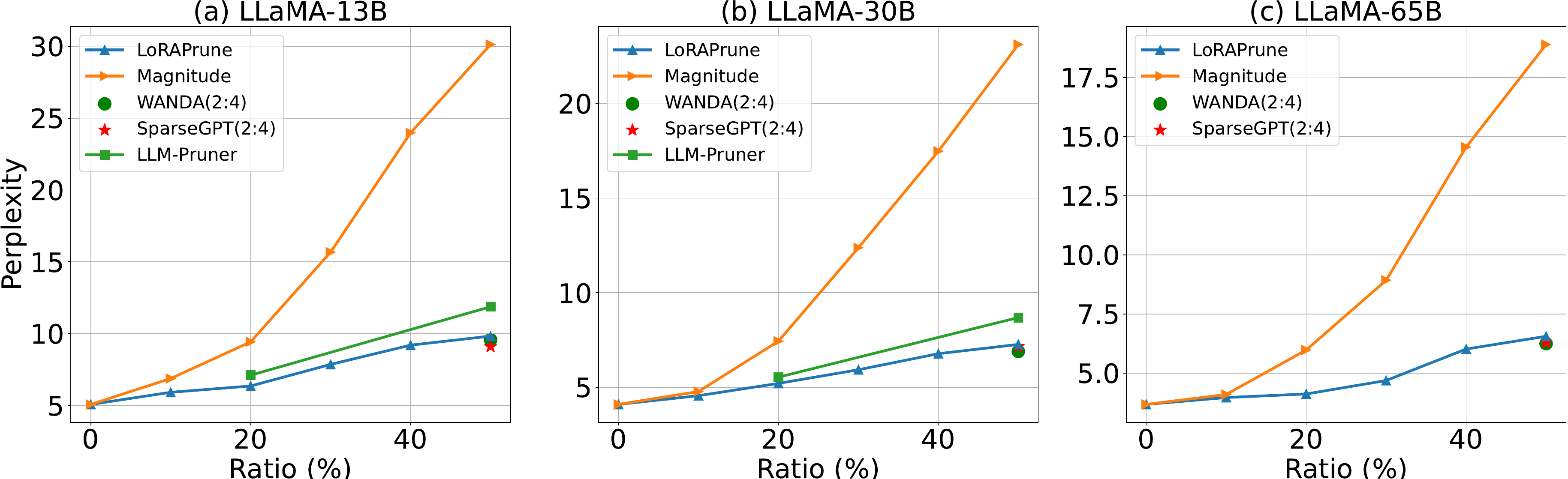}
\end{center}
\caption{Pruning results on large-scale LLMs: (a) LLaMA-13B, (b) LLaMA-30B, (c) LLaMA-65B.}
\label{fig:ratios}
\end{figure*}

\begin{figure*}[h]
\begin{center}
\includegraphics[scale=0.22]{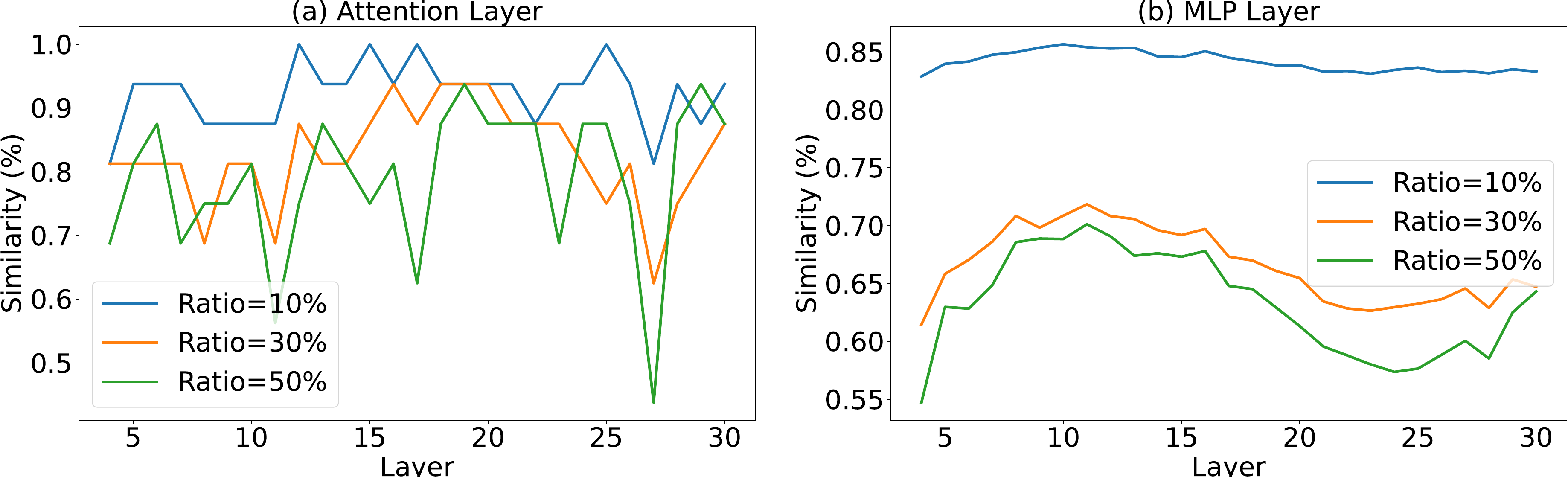}
\end{center}
\caption{Similarity between LoRA gradient and vanilla criterion on (a) Attention, (b) MLP layers.}
\label{fig:sim_lora_taylor}
\end{figure*}

\noindent To this end, we can efficiently and accurately estimate the importance of each group. We then prune the unimportant groups by setting a binary mask $\mathbf{M} \in \{0, 1\} ^{1\times G}$ for each pruned layer. The binary mask $\mathbf{M}$ is 
obtained
by
\begin{equation}
    {\begin{array}{ll}
    \small
    \begin{aligned}
    \mathbf{M}_g =
    \left\{\begin{array}{ll} 
    1 ~~~~~ \bm{\mathcal{\bar{G}}}_g > p \\
    0 ~~~~~ \bm{\mathcal{\bar{G}}}_g \leq p
    \end{array}\right.
    \end{aligned},
    \small
    \end{array}}
\label{eq:mask}
\end{equation}
where 
the index $g\in\{1, \ldots,G\}$
denotes the $g$-th group in the layer, and $p$ represents the threshold of importance. Groups falling below this threshold will be pruned.
After setting the mask, the forward process of each pruned layer can be written as 
\begin{equation}
\label{forward}
    \mathbf{z} = (\mathbf{x}\mathbf{W}_0 + \mathbf{x}\mathbf{BA}) \odot \mathbf{M},
\end{equation}
where $\odot$ denotes Hardamard product and can be calculated by broadcast. The complete algorithm of LoRAPrune is given in Algorithm \ref{alg:LoRAPrune}.
\section{Experiments}
\subsection{Experimental Setup}
\textbf{Models and metrics.} 
Our method is applied to the LLaMA-1 model family \citep{touvron2023llama}, which comprises LLaMA-7B, LLaMA-13B, LLaMA-30B and LLaMA-65B. 
Following \cite{frantar2023massive}, we evaluate models on the perplexity metric with WikiText \citep{merity2016pointer} and PTB \citep{marcus1993building} dataset. To assess the zero-shot ability of LLMs, we follow LLaMA to perform zero-shot task classification on common sense reasoning datasets: PIQA \citep{bisk2020piqa}, HellaSwag \citep{zellers2019hellaswag}, WinoGrande \citep{sakaguchi2021winogrande}, ARC-easy \citep{clark2018think}, ARC-challenge \citep{clark2018think}, OpenbookQA \citep{mihaylov2018can}. We evaluate the in-context learning ability under a 5-shot setting on MMLU \citep{hendrycks2020measuring}.

\noindent \textbf{Implementation details.} We provide results for LoRAPrune as a single-shot method and with post-training recovery fine-tuning. We iteratively prune models on LaMini instruction dataset \citep{lamini-lm} for LLaMA-7b and 20k sampled C4 dataset \citep{raffel2020exploring} for LLaMA-13b, LLaMA-30B and LLaMA-65B. Our training configuration includes a batch size of 128, a learning rate set to 1e-4, and a total of 2 training epochs. As the pre-trained weights remain frozen, there is the option to quantize them into 8-bit values to save memory. All models are optimized by AdamW optimizer \citep{he2020learning} with a cosine learning rate decay.

\noindent \textbf{Contenders.} We compare LoRAPrune with the following pruning methods in both fine-tuning and
without fine-tuning
settings: \textbf{1)} Magnitude Pruning: pruning based on the absolute values of model weights. \textbf{2)} LLM-Pruner \citep{ma2023llm}: pruning using criterion in \equationref{importance_w}. \textbf{3)} WANDA \citep{sun2023simple}: pruning based on the magnitude of input features and pre-trained weights. \textbf{4)} Compresso \cite{guo2023compresso}: pruning based on a set of learnable masks.

\begin{table*}[h]
\caption{Pruning resource required by different pruning criteria. }\label{tab:prune_effect}
\centering
\scalebox{0.75}{
    \begin{tabular}{c|c|cccc}
        \hline
        Model & Pruning criteria & Fine-tuning Throughput $\downarrow$ & GPU Memory $\downarrow$&Total time $\downarrow$ & Perplexity $\downarrow$\\
        \hline
        \multirow{4}{*}{\parbox{2.0cm}{LLaMA-7B \\ (Ratio=50\%)}} & Vanilla & 38.87s/iter (+0.0\%) & 38.6G (+0.0\%) & 5.3 h (+0.0\%)& 11.48 (+0.0\%)\\
        & Magnitude & 13.08s/iter (\textcolor{green}{-66.3\%}) & 16.8G (\textcolor{green}{-56.7\%}) & 1.8 h (\textcolor{green}{-66.04\%})& 17.38 (\textcolor{red}{+52.9\%})\\
        & LoRA-guided & 14.13s/iter (\textcolor{green}{-63.6\%}) & 18.3G (\textcolor{green}{-52.6\%}) &2.0 h (\textcolor{green}{-62.26\%})& 11.60 (\textcolor{red}{+1.0\%})\\
        & LoRA-guided (8-bit) & 15.63s/iter  (\textcolor{green}{-59.8\%}) & 13.8G (\textcolor{green}{-64.2\%}) &2.0 h (\textcolor{green}{-62.26\%})& 12.38 (\textcolor{red}{+9.0\%})\\
        \hline
    \end{tabular}}  
\end{table*}

\subsection{Main Results}
\textbf{Zero-shot performance.} 
Table \ref{tab:LLaMA_result} demonstrates the effectiveness of our proposed method. Our LoRAPrune
far surpasses other large model pruning methods under structured sparsity. For instance, at a 50\% compression rate, LoRAPrune achieves a perplexity of 11.60 on WikiText2, significantly outperforming LLM-Pruner's perplexity of 16.41. We also replicate the experimental results of WANDA under structured pruning scenarios. Our findings reveal that the pruning outcomes achieved by WANDA fell short in comparison to gradient-based pruning methods such as LLM-Pruner and LoRAPrune. This observation underscores the superior performance and effectiveness of gradient-based pruning approaches in our experiments.

\noindent It's worth noting that LoRAPrune's efficient approximation for the gradients of the pre-trained weights allows for 8-bit quantization of those weights, greatly reducing the memory requirements for pruning. Moreover, LoRAPrune demonstrates superior pruning results even when models are quantized to 8 bits. These findings underscore the effectiveness and versatility of LoRAPrune in achieving impressive pruning results across various scenarios and compression rates.

\noindent \textbf{Few-shot performance.}
To verify whether the pruned LLMs retain the in context learning capability, we evaluate on the MMLU with 5-shot. As shown in Table \ref{tab:LLaMA_result}, LoRAPrune consistently achieves a higher score than other pruning methods across all sparsity ratios.  Notably, LoRAPrune achieves performance on par with the unpruned LLaMA-7B model at a 20\% sparsity ratio.

\noindent \textbf{Acceleration for pruned LLMs.} Models with structured pruning can be directly sped up in general GPU devices.
We conducted tests with 2048 tokens, averaging the results over 100 trials. We specifically examined the inference time with and without merging LoRA weights into the pre-trained weights. As shown in Table \ref{tab:speedup}, we observed that when pruning 20\% weights, LLM without merging LoRA has an even slower inference speed than LLM with LoRA merged without pruning. In addition, through structured pruning, the model achieves reductions in inference time of 24.7\% and 49.5\% at compression rates of 20\% and 50\%.

\noindent \textbf{Pruning on large-scale LLMs.} Due to the efficient approximation of the pre-trained weights' gradients, LoRAPrune enables iterative pruning on larger-scale LLMs. 
To ensure that all experiments can be conducted on one GPU, we quantize the pre-trained weights of LLaMA-30b and LLaMA-65b to 8 bits.
The experimental results are shown in Figure \ref{fig:ratios}. We observe that, in comparison to the magnitude-based method, LoRAPrune exhibits significant superiority across various scales. Furthermore, in comparison to the
2:4 sparsity model, LoRAPrune achieves comparable pruning results at a 50\% sparsity rate. However, it is worth noting that the 2:4 sparsity model also faces challenges in direct weight merging with LoRA, resulting in additional computational overhead during 
inference. Besides, accelerating 2:4 sparsity models requires specialized hardware support, such as NVIDIA GPUs based on the Ampere architecture, which significantly increases the deployment constraints for 2:4 sparsity models.


\subsection{Ablation Study}
\textbf{Efficiency of LoRA-guided criterion vs. vanilla criterion.} We conduct a comparative analysis of different pruning criteria with respect to their resource requirements and computational efficiency, including GPU memory and throughput. We adopt the vanilla criterion, as outlined in \equationref{importance_w}, as our baseline. For each forward pass, we set the batch size to 1, and we accumulate this process iteratively until we reach a total of 128 accumulations. To ensure robustness and reliability, we compute and subsequently average the results obtained over a span of 100 steps. The comparison results can be found in Table \ref{tab:prune_effect}. Compared to the vanilla criterion, LoRA-guided and LoRA-guided (8bit) criteria demonstrate a significant reduction in GPU memory usage, saving
52.6\% and 64.2\% of the memory, respectively. Moreover, as the LoRA-guided criterion does not require the computation of original gradients, it achieves a 64.6\% increase in throughput compared to the vanilla criterion with comparable performance, greatly enhancing the speed of the pruning process. 

\noindent \textbf{Efficacy of LoRA-guided criterion vs. vanilla criterion.} 
Since the LoRA-guided criterion in \equationref{eq:final_imp} is an efficient approximation of the vanilla criterion in \equationref{importance_w}, we evaluate the effectiveness of the proposed LoRA-guided criterion by comparing mask similarity with the vanilla criterion.
We randomly sample 128 data and then perform one-shot pruning with both LoRA gradient and vanilla criterion. Figure \ref{fig:sim_lora_taylor} illustrates that in the case of low compression rates (Ratio=10\%), the masks generated by these two criteria exhibit a high degree of consistency. As the compression rates increase, the mask similarity may decrease. However, it is crucial to emphasize that LoRAPrune follows an iterative pruning approach. In each pruning iteration, it only needs to precisely identify the least important weights (about top-5\%), thus ensuring the accurate approximation. Hence, the LoRA-guided criterion can attain results that are on par with those of the vanilla criterion while incurring reduced costs.


\section{Conclusion}
In this paper, we have proposed a method to effectively prune and fine-tune LLMs simultaneously, achieving state-of-the-art efficiency-accuracy trade-offs. Specifically, we have proposed a novel LoRA-guided criterion, for evaluating the parameter importance by only computing the LoRA gradients, which greatly reduces the computational resources required for pruning LLMs. Building upon the proposed criterion, we have presented LoRAPrune, a technique that performs efficient joint pruning and fine-tuning without the need for computing gradients of the pre-trained weights. Finally, comprehensive experiments on various LLMs and benchmarks have demonstrated the superiority of LoRAPrune over other pruning methods. In terms of comparison with the vanilla criterion, the LoRA-guided criterion shows its efficiency and effectiveness. In the future, we aim to further enhance the pruning results of LoRAPrune at higher compression rates.

\noindent\textbf{Limitation.}
LoRAPrune requires fine-tuning to restore model performance. This limitation can restrict the application of LoRAPrune in scenarios where fine-tuning is unavailable.

\noindent{\bf Acknowledgements:}
This work was supported by National Key R\&D Program of China (No. 2022\-ZD\-0\-118\-700), National Natural Science Foundation of China (No.\ 623\-733\-29) and Baima Lake Laboratory Joint Funds of the Zhejiang Provincial Natural Science Foundation of China (No.\ LBMHD\-24\-F03\-0002).
\bibliography{mendeley}
\bibliographystyle{acl_natbib}

\clearpage

\appendix

\label{sec:appendix}

\input{Appendix}

\end{document}

%% file: Appendix.tex
\begin{center}
	{
		\Large{\textbf{Appendix}}
	}
\end{center}
\begin{figure*}[htbp]
\begin{center}
\includegraphics[scale=0.6]{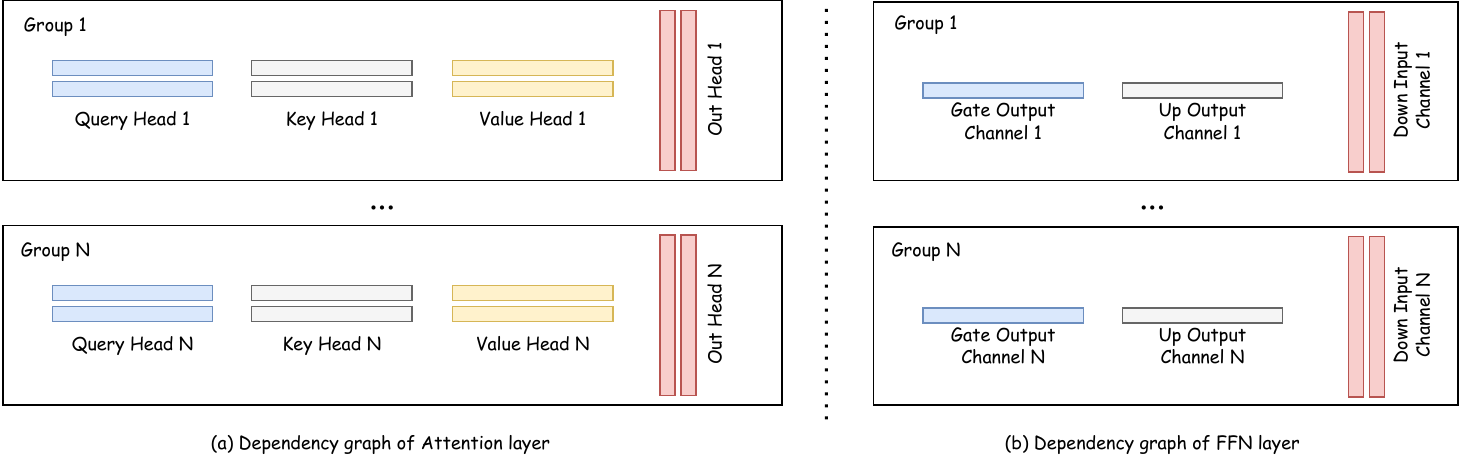}
\end{center}
\caption{Weight dependency in (a) Attention layer, (b) FFN layer.}
\label{fig:dependency}
\end{figure*}
\section{Weight Dependency for LLaMA}
Here, we use LLaMA architecture as an example to explain the weight dependency. The dependency details are shown in Figure \ref{fig:dependency}.
In terms of the Attention module, when we decide to prune a specific head of weights in the \rm{Query} layer, it is imperative that the corresponding weights with the same index in the \rm{Key}, \rm{Value} and \rm{Out} layers are also pruned. Similarly, for the Feed-Forward Network (FFN) module,  when pruning a particular channel of weights in the \rm{Up} layer, it is essential to prune the weights with matching indices in the \rm{Gate} and \rm{Down} layers. This meticulous coordination ensures that pruning maintains the structural integrity and functionality of the model. Following \cite{ma2023llm} and \cite{Fang_2023_CVPR}, we prune heads for \rm{Attention} and channels for \rm{FFN}, respectively. The dependency details are shown in Figure \ref{fig:dependency}.

\section{More Ablation Studies}

\textbf{Pruning on 20k sampled C4 dataset.} 
We also evaluate LoRAPrune on a tiny dataset that randomly samples 20k data from C4 dataset. As presented in Table \ref{tab:LLaMA_c4_result}, LoRAPrune outperforms both LLM-Pruner and WANDA across the majority of zero-shot reasoning datasets, thereby securing the highest average score overall. Specifically, LoRAPrune exceeds the performance of LLM-Pruner by margins of 0.82\% and 1.02\%, respectively.
\begin{table*}[t]
    \centering
    \caption{Zero-shot performance of the compressed LLaMA models fine-tuned on the 20k sampled C4 dataset.
    The average accuracy is calculated among seven classification datasets. \textbf{Bold}/ denotes the best performance at the same compression rate.
    $^{\star}$ denotes the results obtained by our reproduction. 
    } \label{tab:LLaMA_c4_result}
    \vspace{-1em}
    \resizebox{\linewidth}{!}{
    \begin{tabular}{ll|cccccccc}
        \toprule
        \toprule
        Pruning Ratio & Method &  BoolQ & PIQA & HellaSwag & WinoGrande & ARC-e & ARC-c & OBQA & Average$\uparrow$ \\
        \midrule  
        \multirow{1}{*}{Ratio = 0\%} & LLaMA-7B~\citep{touvron2023llama} & 73.18 & 78.35 & 72.99 & 67.01 & 67.45 & 41.38 & 42.40 & 63.25 \\
        \midrule
        \midrule
        \multirow{5}{*}{\parbox{2.0cm}{Ratio = 20\%}} 
        &Magnitude $^{\star}$ & 61.89 & 70.81 & 58.34 & 56.87 & 54.87 & 34.02 & 38.40 & 53.59 \\
        &WANDA$^{\star}$ \citep{sun2023simple}& 65.75 & 74.70 & 64.52 & 59.35 & 60.65 & 36.26 & 39.40 & 57.23 \\
        & LLM-Pruner \citep{ma2023llm} & 64.62 & 77.20 &  68.80 & 63.14 & 64.31 & 36.77 & \textbf{39.80} & 59.23 \\
        &\cellcolor{myblue}LoRAPrune-8bit (Ours) & \cellcolor{myblue}65.37 & \cellcolor{myblue}76.65 &  \cellcolor{myblue}69.41 & \cellcolor{myblue}\textbf{63.78} & \cellcolor{myblue}65.45 & \cellcolor{myblue}36.12 & \cellcolor{myblue}39.50 & \cellcolor{myblue}59.46 \\
        & \cellcolor{myblue}LoRAPrune (Ours)& \cellcolor{myblue}\textbf{65.62} & \cellcolor{myblue}\textbf{79.31} & \cellcolor{myblue}\textbf{70.00} & \cellcolor{myblue}62.76 & \cellcolor{myblue}\textbf{65.87} & \cellcolor{myblue}\textbf{37.69} & \cellcolor{myblue}39.14 & \cellcolor{myblue}\textbf{60.05} \\
        \midrule
        \midrule
        \multirow{5}{*}{\parbox{2.0cm}{Ratio = 50\%}}
        &Magnitude $^{\star}$  &  47.40 & 54.36 & 33.49 & 53.10 & 37.88 & 26.60 & 30.12 & 40.42 \\
        &WANDA $^{\star}$ \citep{sun2023simple} &  50.90 & 57.38 & 38.12 & 55.98 & 42.68 & 34.20 & 38.78 & 45.43 \\
        & LLM-Pruner \citep{ma2023llm} & 60.28 & 69.31 & 47.06 & 53.43 & 45.96 & 29.18 & \textbf{35.60} & 48.69 \\
        &\cellcolor{myblue}LoRAPrune-8bit (Ours) & \cellcolor{myblue}61.43 & \cellcolor{myblue}70.88 & \cellcolor{myblue}47.65 & \cellcolor{myblue}\textbf{55.12} & \cellcolor{myblue}\textbf{45.78} & \cellcolor{myblue}30.50 & \cellcolor{myblue}35.62 & \cellcolor{myblue}49.56 \\
        & \cellcolor{myblue}LoRAPrune (Ours) & \cellcolor{myblue}\textbf{61.88} & \cellcolor{myblue}\textbf{71.53} & \cellcolor{myblue}\textbf{47.86} & \cellcolor{myblue}55.01 & \cellcolor{myblue}45.13 & \cellcolor{myblue}\textbf{31.62} & \cellcolor{myblue}34.98 & \cellcolor{myblue}\textbf{49.71} \\
        \bottomrule
        \bottomrule
    \end{tabular}
    }
\end{table*}

\begin{figure*}[h]
\begin{center}
\includegraphics[scale=0.25]{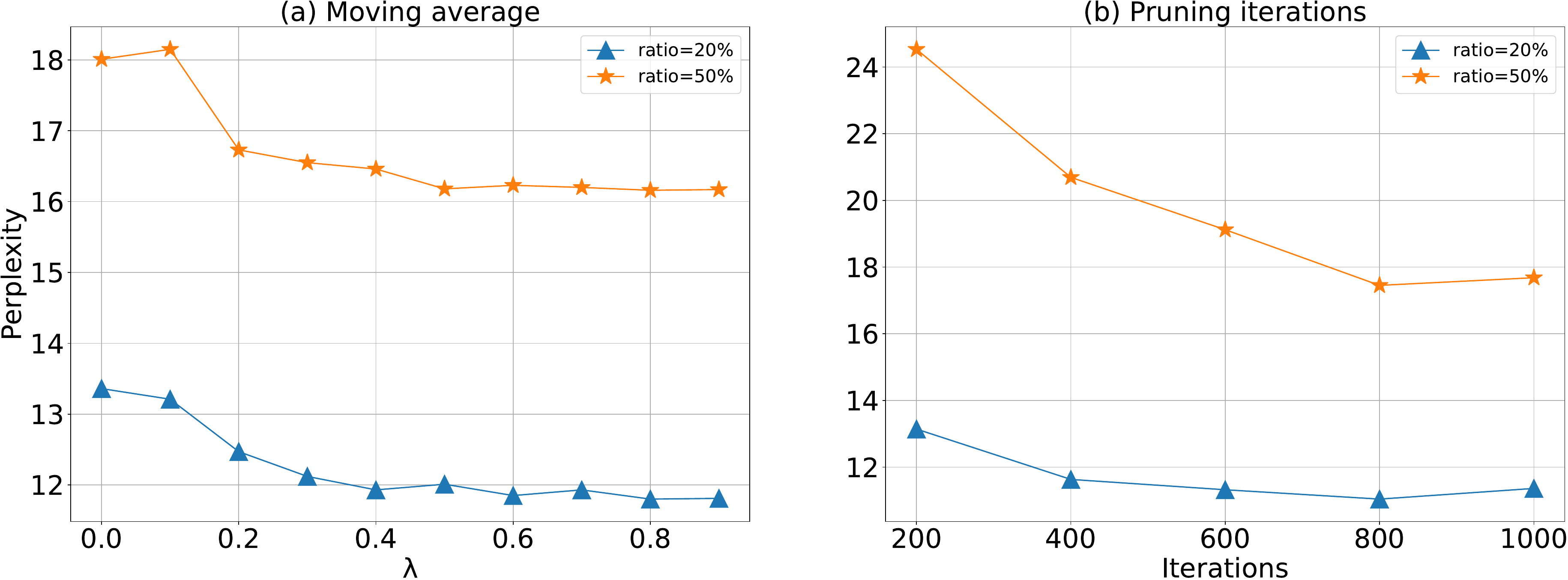}
\end{center}
\vspace{-0.5cm}
\caption{More ablation studies for pruning hyper-parameters: (a) $\lambda$ value in moving average, (b) fine-tuning iterations.}
\label{fig:lambda}
\end{figure*}

\textbf{Effectiveness of the moving average.} 
We verify the rationale behind the moving average through the setting of different values for $\lambda$. These experiments were conducted on LLaMA-7b with 20k sampled C4 dataset. The experimental results, as shown in Figure \ref{fig:lambda} (a), reveal that as $\lambda$ increases, the 
pruning
results exhibit a significant reduction in perplexity. This effect is especially pronounced when $\lambda = 0$ where pruning is solely determined by the importance of the current batch, confirming the effectiveness of the moving average.

\textbf{Impact of iterations.} To assess the impact of the pruning iterations on pruning results, we conducted experiments on the LLaMA-7b model with different iterations on 20k sampled C4 dataset. The results are shown in Figure \ref{fig:lambda} (b), which indicates that excessive iterations can lead to a decrease in the model's zero-shot performance, potentially due to overfitting on the calibration dataset. Furthermore, we observe that the model requires more iterations to regain its performance when pruning with high compression (\eg, ratio=50\%).


\begin{table*}
    \centering
    \caption{Efficiency comparison between LoRAPrune and LLM-Pruner with CPU off-loading.}\label{tab:offloading}
    \vspace{-1em}
    \resizebox{\linewidth}{!}{
    \begin{tabular}{c|cccccc}
        \hline
        Method & Throughput (s/iter) & GPU Memory (GB) & FLOPs (G) &Total time (h)& Pruning time (h)& Fine-tuning time (h)\\
        \hline
        LLM-Pruner & 38.87 &38.6&20298&5.3& 3.5& 1.8\\
        LLM-Pruner + CPU offloading &115.67& 19.5&20298&25.8&24&1.8\\
        \textbf{LoRAPrune (Ours)} &\textbf{14.13} &\textbf{18.3}&\textbf{12881}&\textbf{2.0}&\textbf{0.2}&\textbf{1.8}\\
        \hline
    \end{tabular}
    }
\end{table*}

\textbf{LoRAPrune vs. LLM-Pruner with gradients off-loading.} The gradient off-loading strategy can partially mitigate LLM-Pruner’s memory demands, such as transferring certain gradients to CPU memory. However, the memory access cost and computational overhead are substantial. 
Table \ref{tab:offloading} shows LoRAPrune outperforms LLM-Pruner in efficiency, being 8.19$\times$ faster with CPU offloading and  2.75$\times$ faster without it. This speed allows iterative pruning to counteract the performance drop due to structured sparsity.

\textbf{Joint vs. separate.} 
To demonstrate the necessity of integrating pruning and fine-tuning, we conducted experiments that sequentially performed pruning followed by fine-tuning, specifically applying one-shot pruning to the LLaMA-7b model and then employing LoRA fine-tuning to recover the model's performance.
The experimental results presented in Table \ref{tab:joint} indicate that joint pruning and fine-tuning yields much better performance than the separate counterpart, especially under the high compression ratio. 

\textbf{Pruning frequency.} We explore the impact of different pruning frequencies, \ie, how many iterations of fine-tuning before pruning, on the final performance. The experimental results, as shown in Table \ref{tab:diff_freq}, indicate that our default frequency (frequency=10) obtains the best pruning result.
Additionally, we observe that if pruning is too frequent (frequency=1), the model may not have enough iterations to recover through fine-tuning, leading to inaccurate importance estimation. Furthermore, excessive fine-tuning between pruning iterations (frequency=20) leads to overfitting on the calibration data.


\begin{table*}[t]
  \centering
  \begin{minipage}[b]{0.48\linewidth}
    \centering
    \caption{Effect of the joint pruning and fine-tuning. ``Average'' represents the average performance on seven classification datasets.} \label{tab:joint}
    \vspace{-1em}
    \resizebox{\linewidth}{!}{
    \begin{tabular}{c|c|ccc}
        \hline
        & Method & WikiText2$\downarrow$ & PTB$\downarrow$ & Average$\uparrow$\\
        \hline
        \multirow{2}{*}{Ratio=20\%}
        & Joint & \bf{12.93} & \bf{22.52} & \bf{60.05}\\
        & Separate & 14.51 & 24.30 & 57.18 \\
        \hline
        \multirow{2}{*}{Ratio=50\%}
        & Joint &  \bf{18.37} & \bf{28.68} & \bf{49.71} \\
        & Separate & 21.78 & 40.39 & 47.56\\
        \hline
    \end{tabular}
    }
  \end{minipage}
  \hfill
  \begin{minipage}[b]{0.43\linewidth}
    \centering
    \caption{Results under different pruning frequencies. ``Average'' denotes the average performance on seven classification datasets.} \label{tab:diff_freq}
    \vspace{-1em}
    \resizebox{\linewidth}{!}{
    \begin{tabular}{c|ccc}
        \hline
        Frequency & WikiText2$\downarrow$ & PTB$\downarrow$ & Average$\uparrow$\\
        \hline
        1 & 18.57 & 29.02 &  49.37\\
        5 & 18.63 & 28.73 &  49.16 \\
        10 & \bf{18.21} & \bf{28.68} & 49.59 \\
        20 &  18.47 & 28.70 & \bf{49.71} \\
        \hline
    \end{tabular}
    }
  \end{minipage}
\end{table*}

\section{Generative Results}
\label{apsec:instruction}
We show the generalization capability of the LoRAPrune by some instructions encompass common sense, translation, and coding tasks in Table \ref{tab:gener_results}. 

\begin{table*}[t]
    \centering
    \caption{Generated examples from the Pruned models} 
    \label{tab:gener_results}
    \vspace{-0.5em}
    \resizebox{\linewidth}{!}{
    \tiny
    \begin{tabular}{p{1.0cm}|p{9.8cm}}
        \hline
        Model & Generated Sentences \\
        \midrule
        \multirow{14}{*}{\parbox{0.8cm}{\makecell{LLaMA-7b \\(ratio=0\%)}}} & \underline{Instruction: Tell me about alpacas.} Response: Alpacas are members of the camelid family and are native to the Andes Mountains of South America. They are known for their soft, luxurious fleece, which is used to make clothing, blankets, and other items. Alpacas are herd animals and live in groups of up to 20 individuals. They are generally docile and easy to care for, making them popular as pets and livestock. Alpacas can live up to 20 years in captivity, and their fleece can be sheared up to three times a year. \\
        \cmidrule{2-2}
         & \underline{Instruction: Translate the sentence 'I have no mouth but I must scream' into Spanish.}
Response: I no tengo boca pero tengo que gritar.
 \\
        \cmidrule{2-2}
         & \makecell[l]{\underline{Instruction: Write a Python program that prints the first 10 Fibonacci numbers.} Response: \\
def fibonacci(n): \\
\quad if n == 0:\\
\quad \quad return 1 \\
\quad if n == 1:\\
\quad \quad return 1 \\
\quad else: \\
\quad \quad return fibonacci(n - 1) + fibonacci(n - 2)\\
for n in range(1, 10): \\
\quad print(fibonacci(n)) }\\
        \hline
        \multirow{14}{*}{\parbox{0.8cm}{\makecell{LLaMA-7b \\ (ratio=20\%)}}} & {Instruction: Tell me about alpacas.} Alpacas are a species of camelid native to South America. They are known for their soft, luxurious wool, which is used to make clothing, blankets, and other textiles. Alpacas are also used for their meat, which is considered a delicacy in some parts of the world. Alpacas are social animals that live in herds and are known for their gentle demeanor. They are also known for their intelligence and ability to be trained for various tasks, such as pulling carts or performing tricks for entertainment.  \\
        \cmidrule{2-2}
         & \underline{Instruction: Translate the sentence 'I have no mouth but I must scream' into Spanish.}
Response: Yo no tengo la boca pero tengo que gritar. \\
        \cmidrule{2-2}
         & \makecell[l]{\underline{Instruction: Write a Python program that prints the first 10 Fibonacci numbers.} Response: \\
def fibonacci(n): \\
\quad if n == 0 or n == 1: \\
\quad \quad return 1 \\
\quad else: \\
\quad \quad return fibonacci(n - 1) + fibonacci(n - 2)\\
for n in range(1, 10): \\
\quad print(fibonacci(n)) }\\ 
        \hline
    \end{tabular}
    }
    \vspace{-1em}
\end{table*}